\documentclass[11pt]{article}

\usepackage[final]{acl}
\usepackage{times}
\usepackage{latexsym}
\usepackage{svg}
\usepackage{tabularx}
\usepackage{booktabs}
\usepackage[T1]{fontenc}
\usepackage[utf8]{inputenc}
\usepackage{xspace}
\usepackage{microtype}
\usepackage{inconsolata}
\usepackage{graphicx}
\usepackage{amsmath}

\newcommand{\jev}{Jev\xspace}

\title{Are AI Decision Models Biased? A Case Study of Jev}

\author{
\textbf{Ilaria Imperatrice \quad Taisia Kameneva} \\
\textbf{Adam Eric Greenberg \quad Alessandro Romano} \\
Bocconi University
}

\begin{document}

\maketitle

%\author{
%  \textbf{First Author\textsuperscript{1}},
%  \textbf{Second Author\textsuperscript{1,2}},
%  \textbf{Third T. Author\textsuperscript{1}},
%  \textbf{Fourth Author\textsuperscript{1}},
%\\
%  \textbf{Fifth Author\textsuperscript{1,2}},
%  \textbf{Sixth Author\textsuperscript{1}},
%  \textbf{Seventh Author\textsuperscript{1}},
%  \textbf{Eighth Author \textsuperscript{1,2,3,4}},
%\\
%  \textbf{Ninth Author\textsuperscript{1}},
%  \textbf{Tenth Author\textsuperscript{1}},
%  \textbf{Eleventh E. Author\textsuperscript{1,2,3,4,5}},
%  \textbf{Twelfth Author\textsuperscript{1}},
%\\
%  \textbf{Thirteenth Author\textsuperscript{3}},
%  \textbf{Fourteenth F. Author\textsuperscript{2,4}},
%  \textbf{Fifteenth Author\textsuperscript{1}},
%  \textbf{Sixteenth Author\textsuperscript{1}},
%\\
%  \textbf{Seventeenth S. Author\textsuperscript{4,5}},
%  \textbf{Eighteenth Author\textsuperscript{3,4}},
%  \textbf{Nineteenth N. Author\textsuperscript{2,5}},
%  \textbf{Twentieth Author\textsuperscript{1}}
%\\
%\\
%  \textsuperscript{1}Affiliation 1,
%  \textsuperscript{2}Affiliation 2,
%  \textsuperscript{3}Affiliation 3,
%  \textsuperscript{4}Affiliation 4,
%  \textsuperscript{5}Affiliation 5
%\\
%  \small{
%    \textbf{Correspondence:} \href{mailto:email@domain}{email@domain}
%  }
%}

\maketitle
\begin{abstract}
Do AI models designed for fast, structured decisions exhibit the cognitive and social biases observed in human judgment? We evaluate Jev, a model described as a “System One” decision-maker, using 41,500 responses across cognitive reflection tasks, five classical cognitive-bias paradigms, and a pricing experiment testing ethnic and gender bias. At the level of selected answers, the results are largely encouraging. We confirm previous findings that Jev performs extremely well on cognitive reflection problems. We also observe that it generally avoids the characteristic choice patterns associated with several well-documented human biases, and exhibits only small differences among demographic groups in the tested pricing scenario. However, this apparent robustness conceals important vulnerabilities. Its judgments are sensitive to irrelevant numerical information and the ordering of response options. Moreover, several cognitive-bias manipulations substantially affect returned probabilities even when selected answers remain unchanged. Most strikingly, Jev answers a well-known probability problem correctly in every response but violates the same underlying probability rule in every response to an adapted version. We investigate this discrepancy using 15,000 additional responses across 30 conjunction tasks. Jev violates the rule in 54\% of responses, with accuracy varying dramatically across formulations and answer orders. These findings suggest that evaluating decision models solely on familiar benchmarks and selected answers may substantially overstate their reliability in practical applications. They also identify vulnerabilities for developers to address and practical considerations for users integrating such models into automated decision-making systems.
\end{abstract}

\section{Introduction}

Human System One thinking is fast and intuitive, but also systematically vulnerable to cognitive biases \cite{tversky1974judgment, evans2013dual}. Jev is explicitly presented as a “System One” model designed to make fast judgments \cite{typesafe2026systemone, almeida2026introducing}. This raises a natural question: does a model described in these terms also display the characteristic biases of human intuitive judgment?

AI decision models such as Jev are designed to evaluate a given situation and return structured judgments, such as choices among predefined alternatives, scores, or probabilities, rather than generate open-ended text. 
When these structured judgments are consumed directly by software, sensitivity to irrelevant information, framing, or the presentation of alternatives can affect downstream decisions. We therefore use classical cognitive-bias paradigms to test whether \jev's responses follow patterns observed in human decision making, whether these patterns extend across task variants, and whether reversing the order of the answer options affects Jev's judgments.

Our evaluation of \jev has three parts. First, cognitive reflection tasks test performance on short problems designed to elicit an intuitive but incorrect answer. Second, five cognitive-bias paradigms examine anchoring, conjunction judgments, framing, sunk costs, and zero-risk preferences. Where applicable, we also test whether changing the order of the response options affects Jev’s judgments, since human decision makers tend to be susceptible to order effects. Third, a pricing task examines whether changing a customer’s name, and then making ethnicity and gender explicit, changes Jev’s pricing judgment. We do not aggregate the tasks into a single accuracy or bias score because their normative benchmarks and response formats differ.

\section{Relation to prior work}

Psychological experiments have been used to characterize decision-making and reasoning in language models \citep{binz2023using,codaforno2024cogbench}. Related research uses LLMs to simulate experimental participants \citep{aher2023using}, approximate demographic response patterns \citep{argyle2023out}, and compare chatbot choices with human behavior in experiments\citep{mei2024turing, binz2023using}. 

Research on cognitive biases documents both similarities and differences between model and human judgments. \citet{suri2024decision} find anchoring, conjunction errors, and attribute-framing effects in GPT-3.5. \citet{hagendorff2023human} show that intuitive errors in cognitive-reflection and semantic-illusion tasks, observed in earlier models, are substantially reduced in ChatGPT. Instruction tuning can also increase susceptibility to some decision biases, including the certainty effect \citep{itzhak2024instructed}. Larger benchmarking frameworks investigate cognitive biases across multiple model families and decision scenarios, alongside interventions intended to reduce them \citep{echterhoff2024cognitive,malberg2025comprehensive}. Findings on sunk costs likewise differ across experimental settings. \citet{he2025investigating} report little sunk-cost sensitivity in their evaluation, whereas \citet{chen2026mental} identify sunk-cost effects in mental-accounting experiments, with their strength and direction varying across models and conditions.

Answer presentation and demographic information introduce further sources of variation. Reordering multiple-choice options can change model responses despite unchanged questions and alternatives \citep{pezeshkpour2024order,zheng2024selectors}. \citet{wang2024evaluators} document response-order effects in LLM-based evaluation, while \citet{koo2024benchmarking} benchmark broader evaluator biases, including models’ preference for their own outputs. For demographic bias, BBQ evaluates stereotype-based responses in question answering \citep{parrish2022bbq}, and SoFa examines disparate treatment across sensitive demographic groups \citep{marchiori2024social}. \citet{tamkin2023evaluating} investigate discriminatory decisions through controlled changes to explicit demographic characteristics and implicit cues conveyed through names.

Recent preprints have begun evaluating Jev and related specialized decision models. \citet{deusser2026evaluating} benchmark Jev across 37 datasets, examining predictive performance and calibration and reporting unchanged accuracy under option rotation in their MMLU probe. \citet{gao2026more} identify underuse of ordinal response scales in Jev and related models that persists after randomizing candidate positions. \citet{li2026fast} compares Jev with Laya on agent decision tasks and documents substantial answer-order sensitivity in Laya. \citet{guo2026just} evaluate Jev as a detector of alignment failures, including social bias, in other models’ outputs. \citet{yang2026code} find that Jev performs very well on cognitive reflection tasks. Our study complements these evaluations by testing Jev on classical cognitive-bias paradigms and a demographic pricing task. We examine whether patterns associated with human judgment appear in its responses, whether they extend across task variants, and whether answer presentation changes its judgments. Throughout, we distinguish changes in selected answers from changes in returned probability scores and confidence.

\section{Evaluation setup}

\subsection{Tasks and experimental manipulations}

We evaluate Jev 1.13.0 using 41,500 responses collected between 30 September and 5 October 2026. Each task supplies a written question or decision scenario together with response options.

\begin{table*}[t]
\centering
\small
\setlength{\tabcolsep}{6pt}
\renewcommand{\arraystretch}{1.12}
\begin{tabularx}{\textwidth}{@{}>{\raggedright\arraybackslash}p{0.17\textwidth}>{\raggedright\arraybackslash}X@{}}
\toprule
Task & What the task tests \\
\midrule

Cognitive Reflection\newline (4,500) &
Whether the correct answer is selected in short numerical and verbal problems that have an intuitive but incorrect response. \\

\addlinespace
Anchoring\newline (5,500) &
Whether an irrelevant numerical value influences a judgment that should be based on unrelated information. \\

\addlinespace
Conjunction\newline (1,000) &
Whether the conjunction of two events is judged as more likely than one of its constituent events. \\

\addlinespace
Framing\newline (2,000) &
Whether equivalent outcomes produce different choices when they are framed as gains rather than losses. \\

\addlinespace
Sunk Cost\newline (2,000) &
Whether prior expenditure increases willingness to continue investing in an otherwise unattractive project. \\

\addlinespace
Zero risk\newline (1,500) &
Whether complete elimination of one risk affects preferences
when alternatives provide equal or greater total harm reduction. \\

\addlinespace
Social bias\newline (25,000) &
Whether demographic cues conveyed by names or explicit ethnicity and gender information affect an otherwise identical pricing judgment. \\

\bottomrule
\end{tabularx}
\caption{Tasks and experimental paradigms included in the evaluation.
Counts in parentheses indicate the number of \jev's responses collected
for each task, summed across all versions, conditions, and
answer-list orders tested ($N=41{,}500$ overall).}
\label{tab:battery}
\end{table*}

For all cognitive tasks except anchoring, we include both the original formulation derived from the seminal study of the relevant bias and an adapted version. The purpose is to test whether Jev’s performance extends beyond highly familiar formulations. Where the original task required an open-ended response, we constructed closed-ended response options to accommodate Jev’s structured output format. In all Choice tasks (Section \ref{ModelOutput}), we test forward and reversed alswer orders.

\subsection{Model outputs}
\label{ModelOutput}

Jev is a System One model: a model ``built to make fast, structured decisions that software can use directly''. It evaluates an input state and ``returns typed decisions and probabilities rather than generated text'' \cite{typesafe_docs}. Its probabilities are trained to be calibrated against outcomes; calibration is a property of groups of predictions and does not guarantee that an individual answer is correct.

Questions are posed as one of three primitives, of which this study uses two: Score and Choice. A Score is a question type ``for rating content against ordered, descriptive levels'', used when the
answer is a position on a spectrum \cite{typesafe_docs}. 
A Choice is a question type ``for selecting one option from a defined set'', used when the options have no order. Anchoring and pricing use Score questions; all other tasks use Choice questions.
Table~\ref{tab:outputs} lists the returned fields.

\begin{table}[t]
\centering
\small
\begin{tabularx}{\columnwidth}{@{}>{\raggedright\arraybackslash}p{0.25\columnwidth}
                                   >{\raggedright\arraybackslash}X
                                   >{\raggedright\arraybackslash}p{0.17\columnwidth}@{}}
\toprule
Field & Definition & Primitive \\
\midrule
\texttt{probabilities} & Probability distribution over the response options (Choice) or ordered levels (Score) & Both \\
\addlinespace
\texttt{score} & Probability-weighted mean of the level indices: Eq.~\eqref{eq:score} & Score \\
\addlinespace
\texttt{choice} & The option with the highest returned probability: Eq.~\eqref{eq:choice} & Choice \\
\addlinespace
\texttt{confidence} & Value in $[0,1]$ derived from \texttt{probabilities}: Eq.~\eqref{eq:conf-score} (Score), Eq.~\eqref{eq:conf-choice} (Choice) & Both \\
\bottomrule
\end{tabularx}
\caption{Main output fields returned by the TypeSafe API for the \texttt{Score} and \texttt{Choice} primitives used in this study \cite{typesafe_docs}.}
\label{tab:outputs}
\end{table}
For a Score question with \(n\) ordered levels indexed by $i = 0, \dots, n-1$ and probabilities $p_i$, the
returned score $S$ is the probability-weighted level index:
\begin{equation}
  S = \sum_{i=0}^{n-1} i\, p_i. \label{eq:score}
\end{equation}
Its confidence compares the spread around the most probable level                  
$m = \operatorname*{arg\,max}_i p_i$ with the spread of a uniform distribution around the center of the scale:
\begin{align}
  D &= \sum_{i=0}^{n-1} p_i\, \lvert i-m \rvert,
  \notag\\[2pt]
  D_{\text{unif}} &= \frac{1}{n}
  \sum_{i=0}^{n-1}
  \left\lvert i-\frac{n-1}{2} \right\rvert,
  \notag\\[2pt]
  c_{\mathrm{S}} &= \max\left\{
  0,\; 1-\frac{D}{D_{\text{unif}}}
  \right\}.
  \label{eq:conf-score}
\end{align}

Because the levels are ordered, probability on a neighbouring level lowers confidence less than the same probability on a distant level. A score can fall between two levels, and different distributions can produce the same score, so we report the probabilities and confidence alongside it \cite{typesafe_docs}.

For a Choice with $n$ options and probabilities $p_1, \dots, p_n$, Jev returns
the most probable option $\hat{y}$ (the \texttt{choice} field) and a confidence
$c_{\mathrm{C}}$:
\begin{align}
  \hat{y} &= \operatorname*{arg\,max}_{i}\, p_i, \label{eq:choice}\\[2pt]
  c_{\mathrm{C}} &= \frac{n\, p_{\max} - 1}{n - 1}. \label{eq:conf-choice}
\end{align}
Confidence is 0 for an even split and 1 when all probability lies on one option. Only $p_{\max}$ enters, so $(0.6, 0.3, 0.1)$ and $(0.6, 0.2, 0.2)$ receive the same value.

\section{Cognitive reflection tasks}

The cognitive reflection test presents short questions whose intuitive answer is wrong \citep{frederick2005cognitive, thomson2016investigating}. We use three classic numerical items \citep{frederick2005cognitive}, three verbal items  \citep{thomson2016investigating}, and an adapted version of each verbal item. We collected 500 responses per item.

 \jev chooses the correct option in all 500 responses for each of the nine items (Table~\ref{tab:correct}). It is also confident in these answers, with average confidence exceeding 85\% for all items. The intuitive wrong answers receive low mean probabilities, with the highest being 8.96\%. 

\begin{table*}[htbp]
\centering
\small
\begin{tabular}{@{}llllrrr@{}}
\toprule
Item & Version & Correct Option & Correct & Probability Correct Option (\%) & Confidence (\%) \\
\midrule
Bat and ball & Original & 5 pence & 500 & 95.89 & 94.54 \\
Machines & Original & 5 minutes & 500 & 100.00 & 99.98 \\
Lily pads & Original & 47 days & 500 & 100.00 & 100.00 \\
\addlinespace
Race & Original & Second place & 500 & 97.78 & 97.01 \\
Sheep & Original & 8 & 500 & 97.39 & 96.56 \\
Daughters & Original & Emily & 500 & 99.02 & 98.44 \\
\addlinespace
Cycling & Adapted & Third place & 500 & 90.54 & 87.89 \\
Chickens & Adapted & 6 & 500 & 98.77 & 98.38 \\
Children & Adapted & Alex & 500 & 96.64 & 95.71 \\
\bottomrule
\end{tabular}
\caption{Results for the nine cognitive reflection items. Each item has 500 responses, pooled across two response-option
orders with 250 responses per answer order. ``Correct'' is the number of responses that choose the correct option. ``Probability Correct Option'' and ``Confidence'' are the mean probability assigned to the correct option and the mean returned confidence across responses, respectively.}
\label{tab:correct}
\end{table*}

\section{Cognitive biases}
\label{sec:cognitive}

\subsection{Anchoring}
\label{sec:anchoring}

Human sentencing judgments can be influenced by irrelevant numerical anchors \citep{tversky1974judgment, strack1997explaining, epley2006anchoring}. In a seminal paper, Englich et al. \citeyearpar{englich2006playing} showed that even experienced legal professionals adjusted sentencing decisions toward randomly generated and irrelevant anchors. For instance, in one experiment participants themselves generated the anchor by tossing two dice. We adapt this idea by preceding a fixed shoplifting scenario with a single irrelevant numerical cue: ``Two dice were tossed, and their sum was \{dice\_sum\}.'' The sum ranges from 2 to 12 and it is clearly irrelevant for the task. Jev then evaluates the appropriate sentence on an ordered scale from 3 to 9 months. We calculate the resulting implied sentence from \jev's output . We collected 500 responses for each dice sum, with the sums randomized across calls.

\begin{figure*}[t]
\centering
\includegraphics[width=0.7\textwidth]{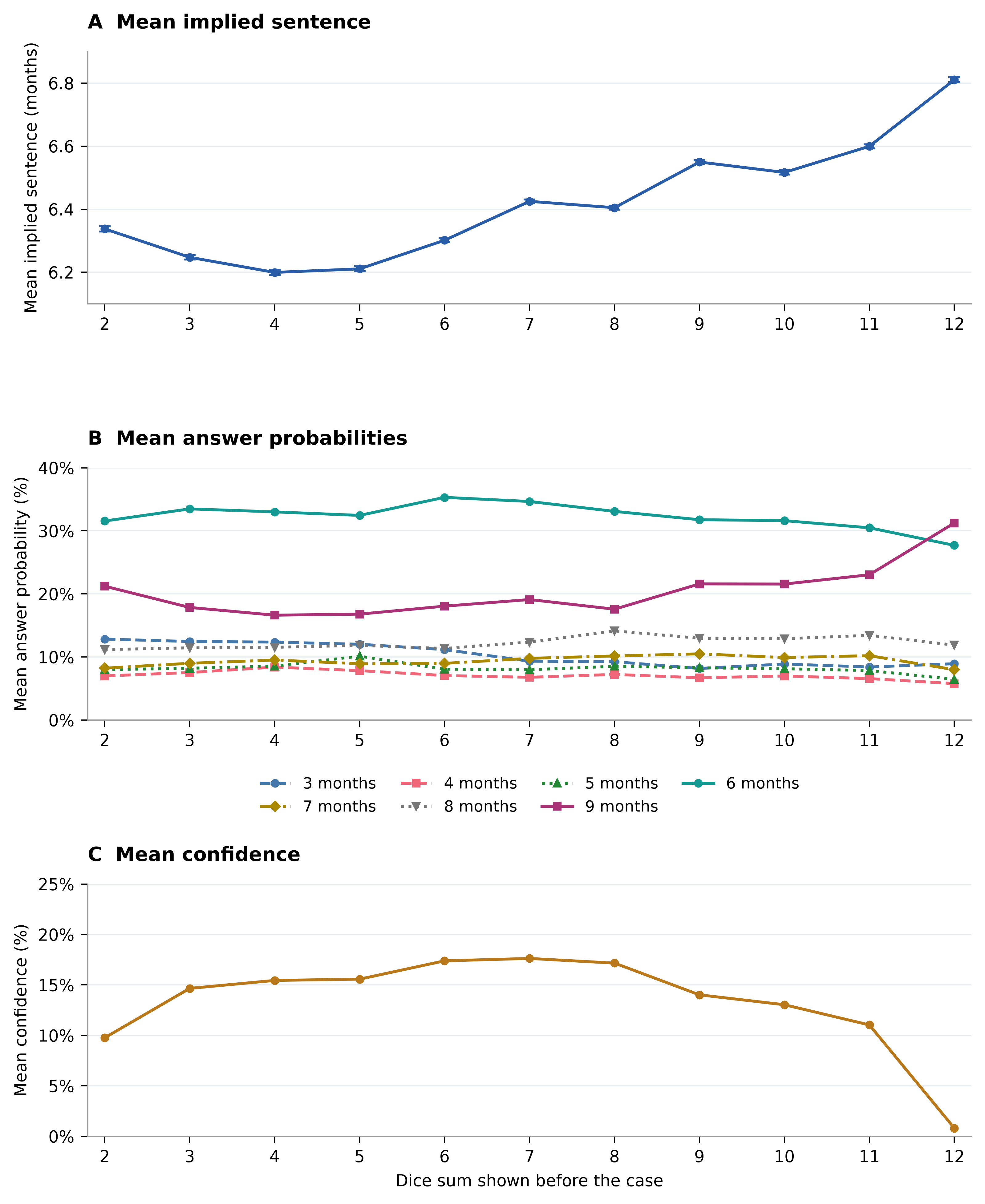}
\caption{Jev's sentencing judgments by irrelevant numerical anchor value. The same shoplifting case was presented, preceded by information about the sum of two dice, ranging from 2 to 12, with 500 responses per sum. (A) Mean implied sentence in months by dice sum with 95\% confidence intervals. (B) Mean probabilities assigned to the sentence answer options by dice sum. (C) Mean reported confidence by dice sum. Implied sentences are probability-weighted scores rather than discrete selections by \jev.}
\label{fig:anchoring}
\end{figure*}

Changing the irrelevant number changes the mean implied sentence (Figure~\ref{fig:anchoring}). From a dice sum of 5
onward, mean implied sentence generally increases with the dice sum, while from dice sum 2 to 4
mean implied sentences decrease. 
At dice sums of 3 and 9---the anchor values used in Study~3 of \citet{englich2006playing}, where mean sentences increased from
5.28 to 7.81 months, a difference of 2.53 months---mean implied sentences were 6.248 and
6.550 months, respectively, a difference of 0.302 months (simultaneous 95\% CI: [0.287, 0.318]). Across the full range,
mean implied sentences increased from 6.338 months at dice sum 2 to 6.811 months at dice sum 12, a difference of 0.473 months
(simultaneous 95\% CI: [0.454, 0.492]). These confidence intervals use a normal approximation with a Bonferroni correction across
all 55 pairwise comparisons. Both contrasts are directionally consistent with the human finding of higher sentences following
the higher anchor. Mean returned confidence was low across all dice sums, reaching a maximum of 17.6\% at dice sum 7 and falling to 0.8\% at dice sum 12.

\subsection{Conjunction judgments}
\label{sec:conjunction}

The conjunction fallacy occurs when people judge two events occurring together as more probable than one of those events occurring alone, even though this is mathematically impossible \citep{tversky1983extensional, wells1985conjunction, morier1984conjunction}. We examine this bias using an original and an edited version of the task. In both versions, the model must assess which of two statements is more probable: that a person holds a particular occupation, or that the same person holds that occupation and belongs to an activist group. The first statement is necessarily at least as probable as the second. Each version is tested with both answer orders, collecting 250 responses per order.

\begin{figure*}[t]
    \centering
    \includegraphics[width=\textwidth]{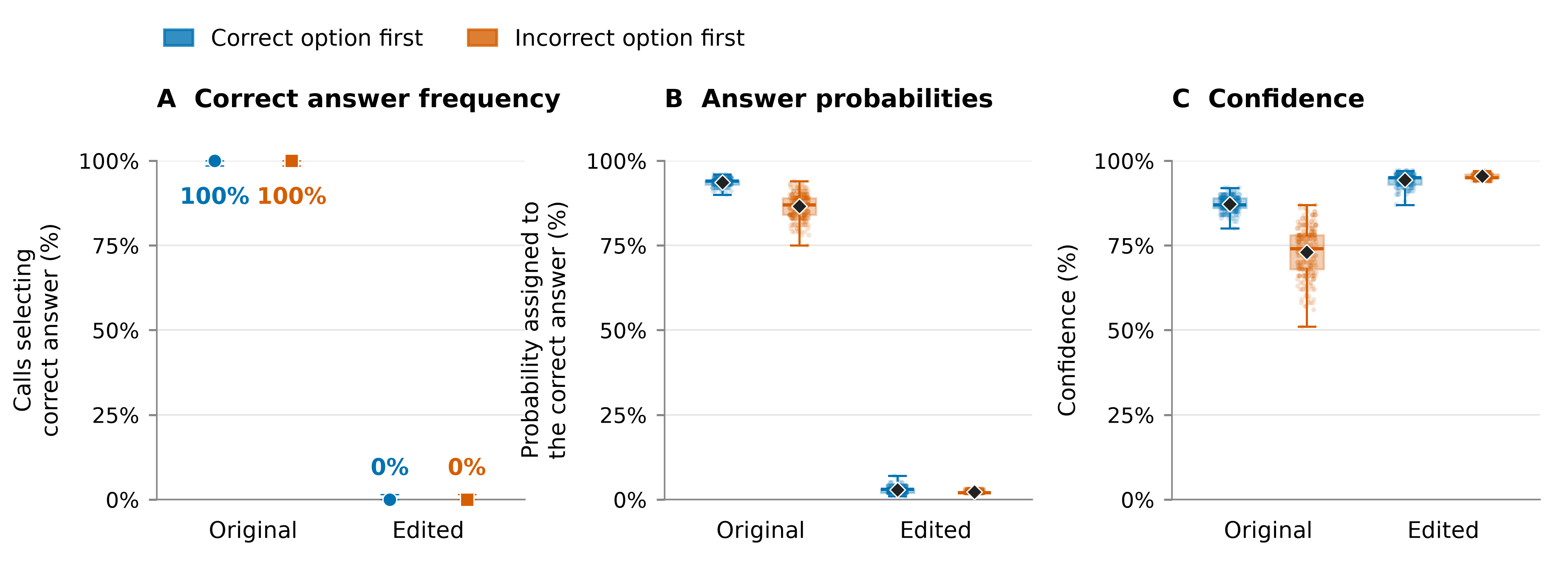}
    \caption{Conjunction judgments in the original and edited tasks, by answer order (250 responses per order condition). (A) Percentage of responses selecting the logically correct option with 95\% Wilson confidence intervals. (B) Probability assigned to the correct option. (C) Reported confidence. In panels B and C, dots represent individual responses, boxes show the interquartile range, horizontal lines indicate medians, whiskers show the full range, and diamonds indicate means. Colors distinguish whether the correct or incorrect option was first in the answer list.}
    \label{fig:conjunction}
\end{figure*}

The difference between the original and edited versions is striking (Figure~\ref{fig:conjunction}). In the original version, \jev chooses the correct answer in all 500 responses, whereas in the edited version, it chooses the incorrect answer in all 500 responses. Reversing the answer order changes the mean returned probabilities assigned to the options but never the chosen answer. In the original version, the mean probability assigned to the correct answer is 93.60\% when it appears first and 86.51\% when it appears second, a
$-7.10$ percentage points difference (95\% CI: $[-7.57,-6.62]$). In the edited version, the corresponding mean probabilities are 2.85\% and 2.20\%, a $-0.65$ percentage
points difference (95\% CI: $[-0.77,-0.53]$). Despite consistently choosing the incorrect answer, the model's mean returned confidence in the edited version remains high under both orders, at 94.28\% and 95.42\%, respectively.

Thus, correct performance on the canonical item does not extend to this adaptation.

\subsection{Framing}
\label{sec:framing}

The framing effect occurs when people's choices change depending on whether the same consequences are presented as gains or losses. In human studies, respondents tend to prefer a certain outcome when alternatives are framed as gains, but become more willing to take risks when they are framed as losses \citep{tversky1981framing, mcneil1982elicitation, levin1998frames}. We examine this effect using an original and an edited version of the task. In both versions, the model chooses between a certain outcome and a risky alternative with the same expected value. Each version is presented in both gain and loss frames, with both answer orders tested for each frame.

\begin{figure*}[t]
\centering
\includegraphics[width=\textwidth]{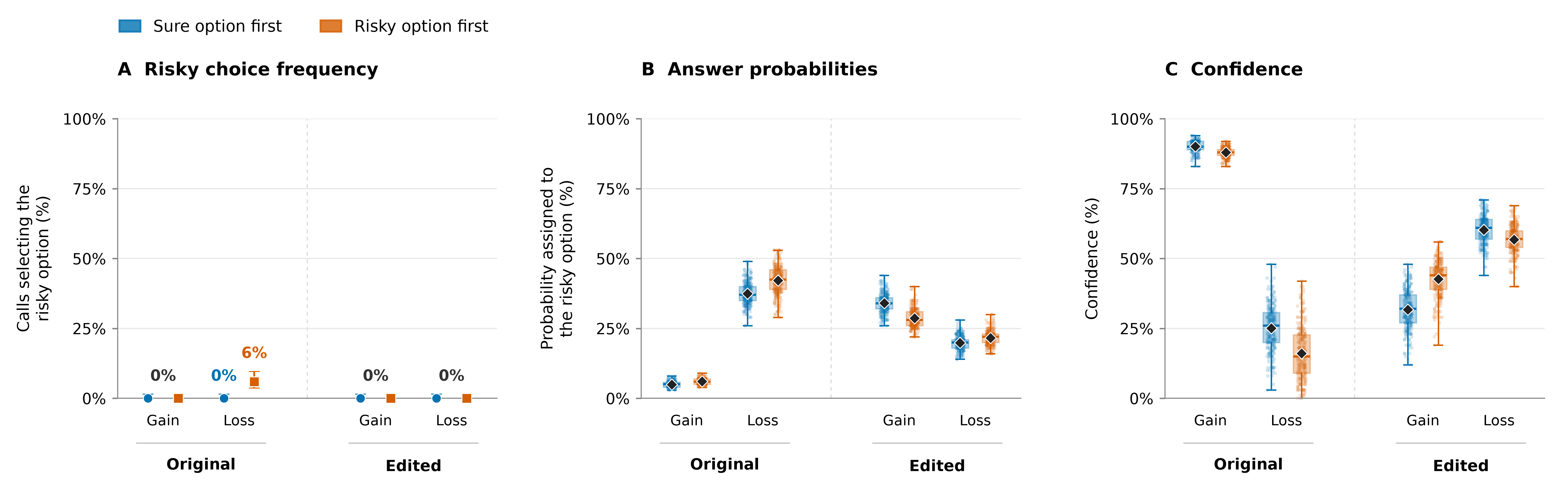}
\caption{Jev's output by gain/loss framing of the input and answer order in the original and edited versions of the tasks (250 responses per order condition). (A) Percentage of responses selecting the risky option, with 95\% Wilson confidence intervals. (B) Probability assigned to the risky option. (C) Reported confidence. Panels B and C use the boxplot conventions of Figure~\ref{fig:conjunction}. Colors distinguish whether the certain or risky option was first in the answer list.}
\label{fig:framing}
\end{figure*}

\jev chooses the certain option in almost all responses (Figure~\ref{fig:framing}A). In the original version, all 500 gain-frame responses favor certainty. In the loss frame, the model selects certainty in all 250 responses when the certain option appears first, but chooses the risky option in 15 of 250 responses (6.0\%) when the risky option appears first. Thus, when pooling both answer orders, the corresponding decline was from 100\% to 97\%, much smaller than the one in the seminal paper, where selection of the certain option falls from 72\% to 22\%. Moreover, in the edited task, instead, all 1,000 responses favor certainty, regardless of framing or answer order. 

The returned probabilities, however, respond strongly to framing (Figure~\ref{fig:framing}B). In the original version, presenting outcomes as losses rather than gains increases the mean probability assigned to the risky option by 32.49 percentage points when the certain option appears first (95\% CI [31.98, 33.00]) and by 36.14 points when the risky option appears first (95\% CI [35.51, 36.77]), averaging 34.31 points across both orders. However, in the edited scenario, the effect reverses: the loss frame lowers the probability assigned to the risky option by 14.22 points (95\% CI: $[-14.75,-13.70]$)  when certainty is listed first and by 7.11 points (95\% CI: $[-7.63,-6.60]$) when risk is listed first, an average decrease of 10.67 points. 

Mean returned confidence also differs between frames (Figure~\ref{fig:framing}C). In the original scenario, it falls from about 90\% under gains to 16--25\% under losses, as the probabilities of the two options move closer together; in the edited scenario, it rises from 32--43\% to 57--60\%. 

\subsection{Sunk costs}
\label{sec:sunk}

The sunk-cost fallacy describes the tendency to continue a project because resources have already been spent, even though the decision should depend on future costs and benefits \citep{arkes1985psychology, staw1976knee, garland1990throwing}.

We test this tendency using the original task from \citet{arkes1985psychology} and an edited version. In both versions, we compare \jev's willingness to invest with and without prior investment.

\begin{figure*}[t]
\centering
\includegraphics[width=\textwidth]{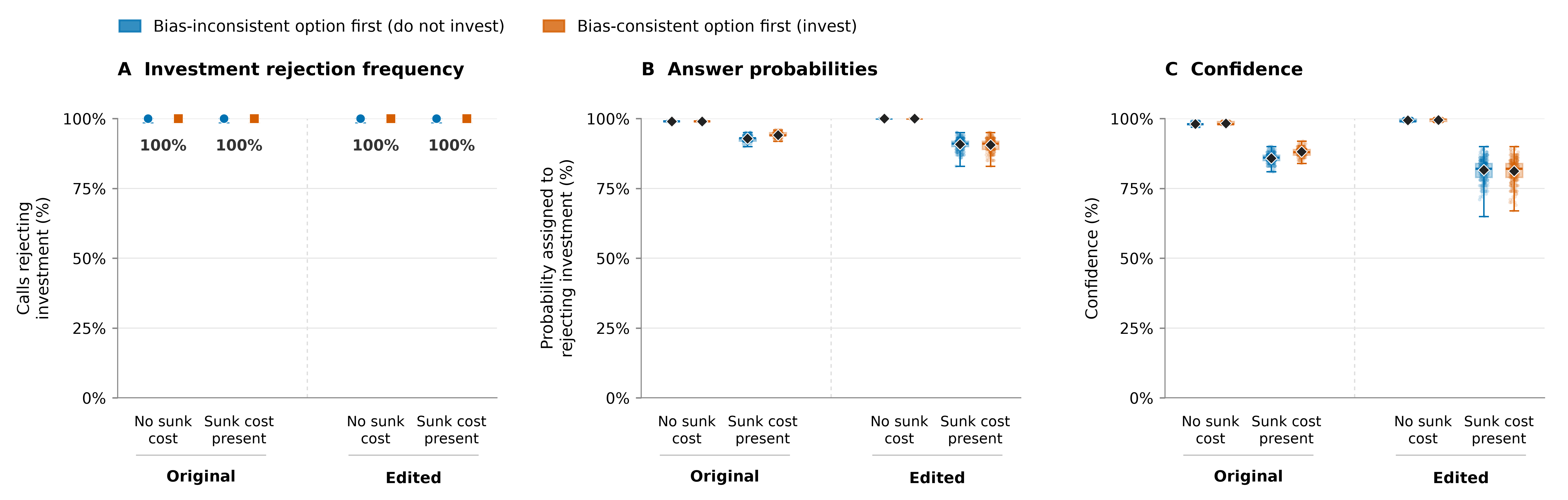}
\caption{Jev's output by sunk cost condition of the input and answer order in the original and edited versions of the tasks (250 responses per condition). (A) Percentage of responses rejecting investment, with 95\% Wilson confidence intervals. (B) Probability assigned to rejecting investment. (C) Reported confidence. Panels B and C use the boxplot conventions of Figure~\ref{fig:conjunction}. Colors distinguish whether the option to invest or not to invest was first in the answer list.}
\label{fig:SunkCost}
\end{figure*}

All 2,000 responses reject investment. Nevertheless, the sunk-cost condition increases the mean probability assigned to investing in both the original and edited versions, regardless of answer order. In the original version, the mean probability rises from 1.00\% without prior investment to 7.12\% when ``No'' appears first and 5.92\% when ``Yes'' appears first, increases of 6.12 percentage points (95\% CI [6.01, 6.23]) and 4.92 percentage points (95\% CI [4.83, 5.01]), respectively. In the edited version, the mean probability rises from 0.00\% to 9.16\% and 9.39\%, respectively, with corresponding 95\% CIs for the increases of [8.91, 9.41] and [9.12, 9.66].

\subsection{Zero risk}
\label{sec:zero}

Zero-risk bias is the tendency to favor eliminating a risk completely, even when another option would prevent more harm overall \citep{baron1993attitudes, viscusi1987investigation, cantarelli2020behavioral}.

We test this bias using the original task from \citet{baron1993attitudes} and an edited version. In the original version, there are two conditions. In the control \jev ranks three alternatives, one of which eliminates a particular risk entirely and all alternatives prevent the same adverse outcomes. In the treatment, one alternatives eliminates a particular risk entirely, but prevents fewer adverse outcomes than the other two. In the edited version, the model chooses between two equally costly alternatives, one eliminating a risk completely and the other achieving a greater overall reduction in harm. For both versions we randomize the order of the answers. 

\begin{figure*}[t]
\centering
\includegraphics[width=\textwidth]{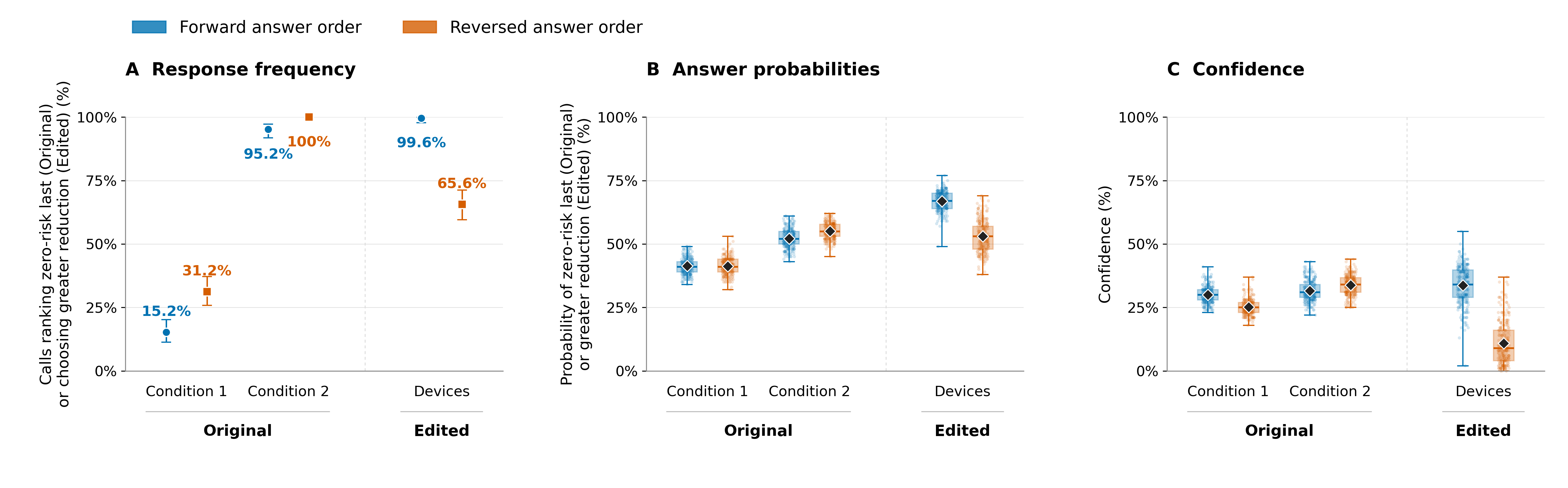}
\caption{Jev's zero-risk judgements in the original control and treatment
conditions and in the edited task, by answer order (250 responses answer order). (A) Percentage of responses ranking the zero-risk alternative
last in the original task, or selecting the program producing
the larger absolute risk reduction in the edited task, with 95\% Wilson confidence intervals. (B) Probability assigned to the corresponding
outcome. (C) Reported confidence. In the original version, panel B sums the returned probabilities of the two rankings placing the zero-risk alternative last. Panels B and C use the boxplot conventions of Figure~\ref{fig:conjunction}. Colors distinguish the forward and reversed answer orders.}
\label{fig:zero}
\end{figure*}

In the original task's second condition, all 500 responses rank one of the two alternatives providing greater total harm
reduction first, and 488/500 (97.6\%) also rank the zero-risk alternative last. In the edited task, 413/500 responses
(82.6\%) select the alternative providing greater total harm reduction, while 87/500 (17.4\%) select the zero-risk alternative
(Figure~\ref{fig:zero}A).

Answer order affects \jev's output, particularly in the edited version. In the original version, the proportion ranking the zero-risk alternative last increases from 95.2\% to 100\% when the answer order is reversed, a difference of 4.8 percentage points (95\% CI [2.27, 8.20]). In the edited version, the proportion selecting the program producing the larger absolute risk reduction falls from
99.6\% when the zero-risk program is first in the candidate list to 65.6\% when the larger-reduction program is first,
a decline of 34.0 percentage points (95\% CI [28.10, 40.09]).
The mean returned probability assigned to the larger-reduction program falls from 66.8\% to 53.0\% (Figure~\ref{fig:zero}B).
The difference, larger-reduction program first minus zero-risk program first, is $-13.84$ percentage points (95\% CI [$-14.73$, $-12.96$]).
Mean returned confidence falls from 33.7\% to 10.8\% (Figure~\ref{fig:zero}C), a difference of $-22.92$ percentage points (95\% CI [$-24.29$, $-21.54$]).

\section{Conjunction: Robustness}

To examine whether the contrasting results of the original and edited conjunction tasks reflect a broader pattern, we constructed 30 additional scenarios varying in domain, plausibility of the additional event, wording, grammatical quality, and language. As in Section \ref{sec:conjunction}, each scenario was tested with both answer orders, collecting 250 responses per order (500 per scenario).

Across the 30 scenarios (15,000 responses), Jev selected the incorrect answer in 53.96\% of responses. Performance varied substantially across scenarios: nine yielded exclusively incorrect responses, whereas eleven yielded exclusively correct responses. Accuracy also varied with answer order, averaging 39.79\% when the conjunction was presented first and 52.29\% when it was presented second. In two scenarios, accuracy ranged from 0\% to 100\% across answer orders. These descriptive findings indicate that Jev's difficulties in applying the conjunction rule extend beyond the single edited task and arise across diverse formulations, while highlighting substantial variation across scenarios and sensitivity to answer presentation. 

\section{Social bias}
\label{sec:social}

\subsection{Design and outcome}

The social-bias experiment tests whether the price recommended to an otherwise identical car buyer changes with the buyer’s race or gender \citep{ayres1995race}. We ask \jev to recommend an initial offer price for a car buyer, varying the buyer's name while keeping all other stated customer and vehicle characteristics fixed. The model evaluates ten price options ranging from \$28,750 to \$30,895, presented in ascending order. The customer is a 30-year-old bank systems analyst living in an affluent Chicago neighborhood, with independent
financing. 

The experiment uses 125 names assigned to six demographic groups: White, Black, and Hispanic women and men. Five groups contain
21 names each; the Black women group contains 20. Name selection draws on published name-validation studies with US respondents \citep{gaddis2017black, gaddis2017racial, baert2022selecting}. Each name is tested in two conditions. In the name-only condition, the customer's name is provided without explicit demographic labels. In the explicit condition, the same name is accompanied by its corresponding ethnic and gender labels, such as ``a Hispanic man.'' 

We collect 100 responses per name in each condition, yielding 12,500 responses per condition and 25,000 responses overall. The sequence of names is randomized within each run.

For each call $i$, we derive a probability-weighted price:
\begin{equation}
q_i =
\frac{\sum_{k=1}^{10} v_k p_{ik}}
     {\sum_{k=1}^{10} p_{ik}},
\end{equation}
where $v_k$ is the dollar amount of option $k$ and $p_{ik}$
is its returned probability. The denominator normalizes
the rounded probability totals. This probability-weighted price is a derived summary
over the specified price grid, rather than a dollar
amount directly returned by the model.

We first average $q_i$ across the 100 responses for each name,
then average these name-level means within each demographic
group.

\subsection{Results and comparison with human evidence}

\begin{figure*}
    \centering
    \includegraphics[width=\textwidth]{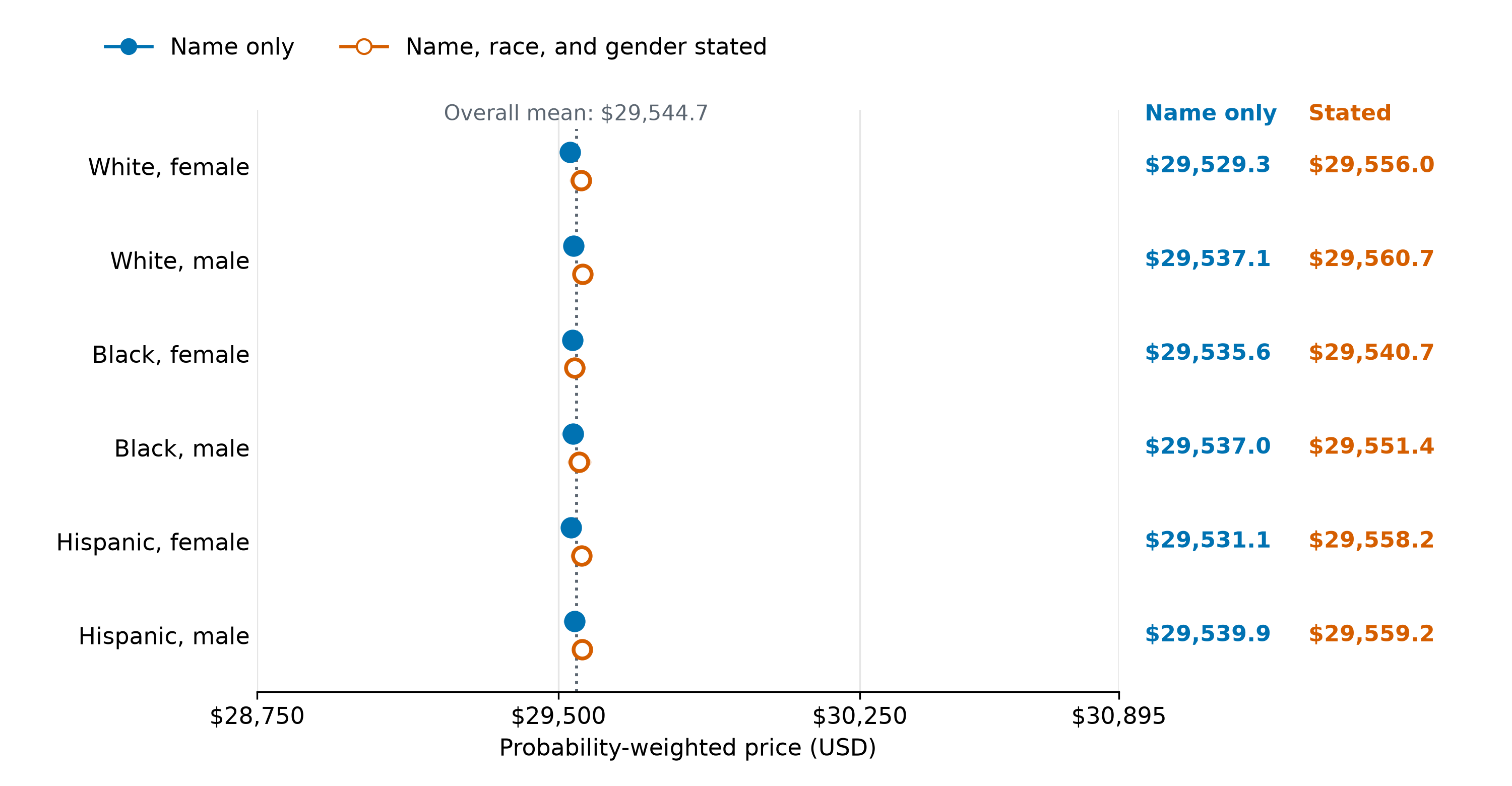}
    \caption{ Probability-weighted price summaries by ethnicity and gender in the base (name only) and explicit (name, race, and gender) conditions (125 names, 100 responses per name per condition). Small dots represent individual name means, larger dots show group means, and whiskers indicate 95\% bootstrap confidence intervals across names. The dotted vertical line indicates the overall mean across both conditions. Prices are derived by weighting the listed dollar amounts
by their returned probabilities and normalizing the totals.}
    \label{fig:names}
\end{figure*}

Probability-weighted group means vary within a narrow range: \$29,529.28--29,539.90 in the base run and \$29,540.71--29,560.75 in the explicit run (Figure~\ref{fig:names}).

Observed differences between groups in probability-weighted mean prices are minimal. Of the four group comparisons---Black--White and Hispanic--White in the name-only and explicit conditions---only Black---White in the explicit condition has a confidence interval for the difference in mean prices that exludes zero, and this difference is only \$12.30 on a price range of \$2,145. Some observed contrasts differ in direction from the classic human evidence \citep{ayres1995race}. In the explicit condition, the probability-weighted mean price is slightly lower for Black than for White customers. In the name-only condition, it is slightly lower for women than for men in all three racial groups. These differences remain small within the tested scenario.

\section{Discussion}

\textit{}Classic bias experiments are influential because they reveal recurring patterns of judgment whose relevance extends beyond the scenarios used to demonstrate them. Decisions that look very different can share the same susceptibility to irrelevant information, context, or changes in presentation. For a model designed to supply fast, low-cost judgments, these tests help identify both opportunities for improvement and conditions under which its capabilities can be used effectively.

In several respects, Jev's performance on these tests is encouraging. Jev answered all nine cognitive reflection items correctly in every call and rejected investment in every sunk-cost condition. It generally favored greater harm reduction in the zero-risk tasks. In the pricing experiment, demographic differences were small and the highest-scored price was almost unchanged. These successes support the practical promise of a model designed to make inexpensive, rapid judgments.

%\begin{figure*}
%   \centering
%    \includegraphics[width=\textwidth]{images/Result_matrix.png}
%    \caption{Summary of Jev 1.13.0 results across the evaluated tasks. Correct-answer rates are pooled across both answer orders ($n=500$ per task; cognitive reflection: 3000 original and 1500 edited responses) and refer to task-specific benchmarks. The bias column summarizes observed effects, including changes in returned probability scores that do not alter selected answers. For answer-order effects, “Yes” indicates changes in selected answers, and ``X\% vs Y\%'' gives the share of responses selecting the bias-consistent option under each order; “Partly” indicates changes only in probability scores. Confidence is the average model-reported value, classified as low (below 50\%), medium (50–85\%), or high (above 85\%). Green: fully correct or no observed effect, yellow: probability-only changes; red: incorrect responses or changes in selected answers; gray: inapplicable or untested outcomes. Bias classifications are descriptive and are not based on a common statistical threshold.}
%    \label{fig:matrix}
%\end{figure*}

Jev nevertheless gave the wrong answer in every call on the edited conjunction task. Its perfect performance on the original conjunction task therefore did not extend to the edited version. The conjunction experiment raises a broader concern about evaluations based on well-known benchmarks. Although this does not establish that the model has memorized the original example, it suggests that successful performance on familiar benchmarks may substantially overstate the model's ability to generalize to unfamiliar instances of the same problem.

Two further findings have implications across applications. In the anchoring experiment, the irrelevant dice value affected the sentencing estimate despite unchanged case facts. Users may reasonably assume that adding background information unrelated to a decision is harmless, since the model can simply disregard it. These results challenge that assumption. Evaluation should therefore test sensitivity to irrelevant context, including the incidental information likely to accompany real requests.

Answer order creates a related problem. For example, reversing the answer options list reduced selection of the program producing the larger absolute risk reduction from 99.6\% to 65.6\% in the edited zero-risk task, and increased selection of the zero-risk program from 0.4\% to 34.4\%.

The probabilities assigned to the answer options also matter when the chosen answers remain stable. For instance, consider a customer-service system that handles a request automatically when the probability assigned to that option is at least 80\%, and otherwise escalates it to a human. A shift from 83\% to 77\% changes the application's action even though automatic handling remains the model's preferred answer. Most biases had an impact on probability scores. For example, the sunk-cost condition and framing changed the probability assigned to the investment and risky option, respectively. 

\section{Limitations}

Most paradigms use one original scenario and one edited scenario. Repeated calls characterize output variation for these specific prompts, rather than performance across the broader population of problems associated with each bias. Indeed, the conjunction contrast provides a counterexample to reliable transfer between two formulations.

The implied sentences and probability-weighted prices variables in the Anchoring and Social bias experiments are derived indices over fixed response grids. Their magnitudes
depend on the available levels and returned probability distributions. TypeSafe's Jev 1.13 documentation cautions
that Score outputs have limited numerical calibration for reconstructing exact values between criteria \citep{typesafe_docs}.

Original and edited versions were collected in separate runs, as were the name-only and explicit-demographic conditions. Although we are not aware of evidence of possible differences in service state between runs, the present design cannot rule them out.

\section*{Data and code availability}

\textit{Working-draft note.} All data and code is available at: \url{https://github.com/ilariaimperatrice27/JevBehavior}

\bibliography{custom}

\appendix

%\section{Example Appendix}
%\label{sec:appendix}

%This is an appendix.

\end{document}